\documentclass{article}
\PassOptionsToPackage{numbers, compress}{natbib}

\usepackage[final]{neurips_2023}

\usepackage[utf8]{inputenc} 
\usepackage[T1]{fontenc}
\usepackage{hyperref} 
\usepackage{url}  
\usepackage{booktabs}
\usepackage{amsfonts}  
\usepackage{nicefrac} 
\usepackage{microtype}  
\usepackage{xcolor}  
\usepackage{hyperref}
\usepackage{url}
\usepackage{caption}
\usepackage{makecell}
\usepackage{graphicx}
\usepackage{blindtext}
\usepackage{bm}
\usepackage{bbm} 
\usepackage{comment}
\usepackage{booktabs}
\usepackage{algorithm}
\usepackage{algorithmic}
\usepackage{multirow}
\usepackage{array}
\usepackage{enumitem}
\usepackage{setspace}
\usepackage[safe]{tipa}
\usepackage{color, colortbl}
\usepackage[misc,geometry]{ifsym}
\usepackage{pifont,makecell,subcaption}
\usepackage{wrapfig}
\usepackage{amsthm}
\usepackage{appendix}
\usepackage{amsmath,amsfonts,bm}

\newcommand{\p}{\mathbf{p}}
\newcommand{\z}{\mathbf{z}}
\newcommand{\q}{\mathbf{q}}
\newcommand{\X}{\mathbf{X}}
\newcommand{\W}{\mathbf{W}}
\newcommand{\Z}{\mathbf{Z}}
\newcommand{\Q}{\mathbf{Q}}
\newcommand{\PP}{\mathbf{P}}

\newcommand{\x}{\mathbf{x}}

\newcommand{\cH}{\mathcal{H}}

\newcommand{\cA}{\mathcal{A}}
\newcommand{\cI}{\mathcal{I}}
\newcommand{\cG}{\mathcal{G}}
\newcommand{\cT}{\mathcal{T}}
\def\Setup{{\textsc{GMI}}}

\theoremstyle{definition}
\newtheorem{exmp}{Example}[section]
\newtheorem*{remark}{Remark}

\title{You Only Submit One Image to Find the Most Suitable Generative Model}

\author{%
  Zhi~Zhou\\
  Nanjing University\\
  \texttt{zhouz@lamda.nju.edu.cn} \\
  \And
  Lan-Zhe Guo\footnotemark[1]\\
  Nanjing University \\
  \texttt{guolz@nju.edu.cn} \\
  \AND
  Peng-Xiao Song \\
  Nanjing University \\
  \texttt{songpx@lamda.nju.edu.cn} \\
  \And
  Yu-Feng Li\thanks{Corresponding Author}   \\
  Nanjing University \\
  \texttt{liyf@nju.edu.cn}
}

\begin{document}
\maketitle
\begin{abstract}
Deep generative models have achieved promising results in image generation, and various generative model hubs, e.g., Hugging Face and Civitai, have been developed that enable model developers to upload models and users to download models. However, these model hubs lack advanced model management and identification mechanisms, resulting in users only searching for models through text matching, download sorting, etc., making it difficult to efficiently find the model that best meets user requirements. In this paper, we propose a novel setting called \emph{Generative Model Identification} (\Setup), which aims to enable the user to identify the most appropriate generative model(s) for the user's requirements from a large number of candidate models efficiently. To our best knowledge, it has not been studied yet. In this paper, we introduce a comprehensive solution consisting of three pivotal modules: a weighted Reduced Kernel Mean Embedding (RKME) framework for capturing the generated image distribution and the relationship between images and prompts, a pre-trained vision-language model aimed at addressing dimensionality challenges, and an image interrogator designed to tackle cross-modality issues. Extensive empirical results demonstrate the proposal is both efficient and effective. For example, users only need to submit a single example image to describe their requirements, and the model platform can achieve an average top-4 identification accuracy of more than 80\%.
\end{abstract}

\section{Introduction}
\begin{figure}[t]
\begin{minipage}{0.58\textwidth}
    \centering
    \resizebox{\linewidth}{!}{
    \includegraphics{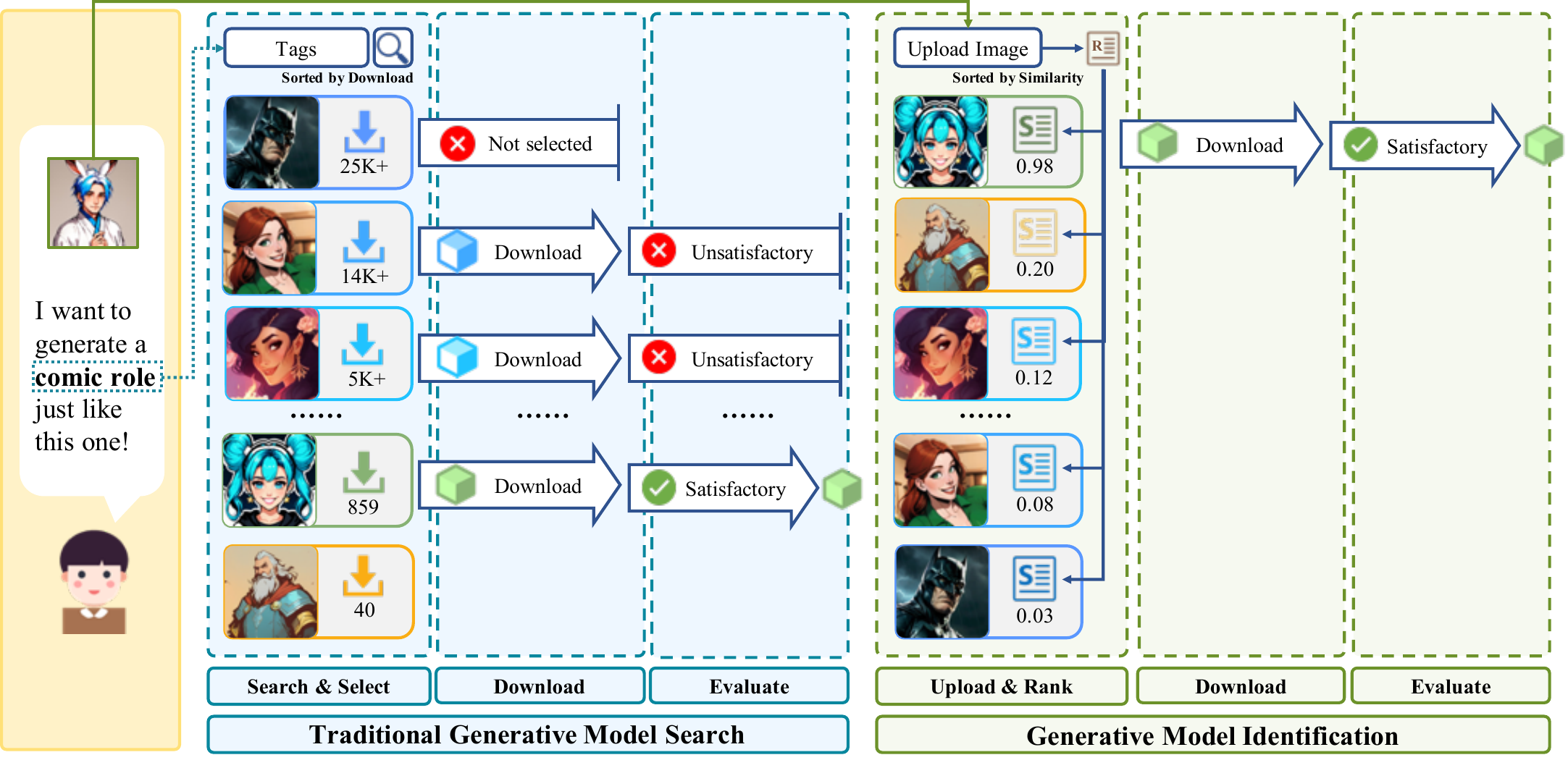}
    }
    \caption{Comparison between traditional generative model search of existing model hubs and \Setup. \Setup\ matches requirements and specifications during the identification process.
    }
    \label{fig:setting}
\end{minipage}
\hfill
\begin{minipage}{0.40\linewidth}
  \centering
  \includegraphics[width=\textwidth]{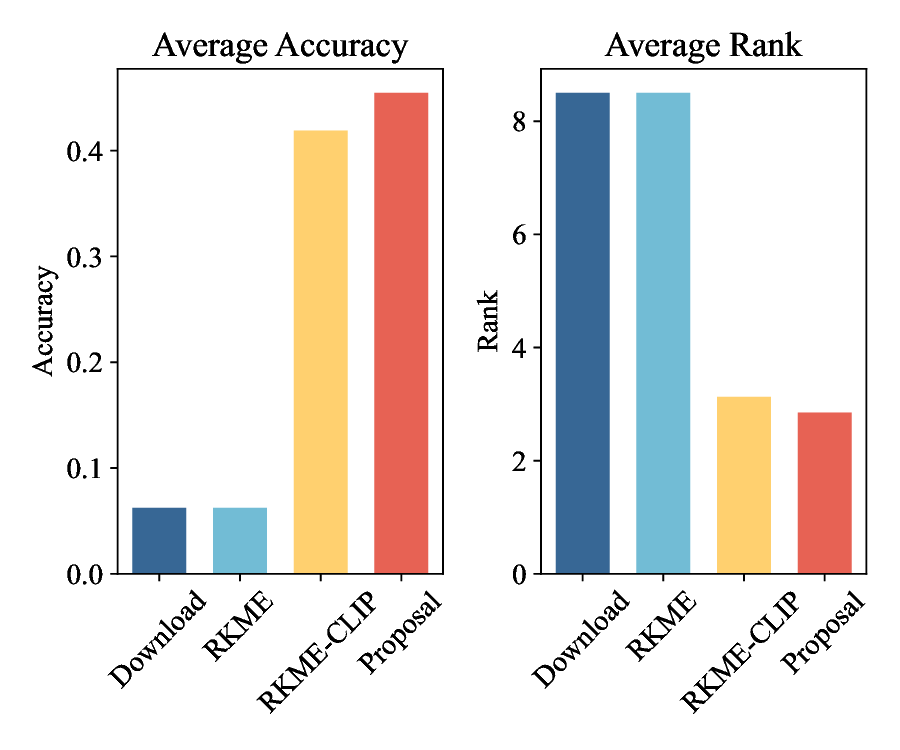}
  \caption{Performance evaluated by average accuracy and rank metrics.}
  \label{fig:performance}
  \end{minipage}

\end{figure}

Recently, stable diffusion models~\cite{SD-DhariwalN21,SD-Rombach22,SD-DicksteinW15,DM-NicholD21} have achieved state-of-the-art performance in image generation and become one of the popular topics in artificial intelligence. Various model hubs, e.g., Hugging Face and Civitai, have been developed to enable model developers to upload and share their generative models. 
However, existing model hubs provide trivial methods such as tag filtering, text matching, and download volume ranking~\cite{GPT-abs-2303-17580}, to help users search for models. However, these methods cannot accurately capture the users' requirements, making it difficult to efficiently identify the most appropriate model for users. As shown in \autoref{fig:setting}, the user should submit their requirements to the model hub and subsequently, they must download and evaluate the searched model one by one until they find the satisfactory one, causing significant time and computing resources.

The above limitation of existing generative model hubs inspires us to consider the following question: Can we describe the functionalities and utilities of different generative models more precisely in some format that enables the model can be efficiently and accurately identified in the future by matching the models' functionalities with users' requirements? We call this novel setting \emph{Generative Model Identification} (\Setup). To the best of our knowledge, this problem has not been studied yet.

It is evident that two problems need to be addressed to achieve GMI, the first is how to describe the functionalities of different generative models, and the second is how to match the user requirements with the models' functionalities. 
Inspired by the learnware paradigm~\cite{Lw-Zhou16}, which proposes to assign a specification to each model that reflects the model's utilities, we enhance the Reduced Kernel Mean Embedding (RKME)~\cite{KME-SriperumbudurFL11, LW-Wu23} to tackle the intractability of modeling generative tasks instead of classification tasks.
To this end, we propose a novel systematic solution consisting of three pivotal modules: a weighted RKME framework for capturing not only the generated image distribution but also the relationship between images and prompts, a pre-trained vision-language model aimed at addressing dimensionality challenges, and an image interrogator designed to tackle cross-modality issues. For the second problem, we assume the user can present one image as an example to describe the requirements, and then we can match the model specification with the example image to compute how well each candidate generative model matches users' requirements. \autoref{fig:setting} provides a comparison between previous model search methods and the new solution. The goal is to identify the most suitable generative model with only one image as an example to describe the user's requirements.

To evaluate the effectiveness of our proposal, we construct a benchmark platform consisting of 16 tasks specifically designed for GMI using stable diffusion models. The experiment results show that our proposal is both efficient and effective. For example, users only need to submit one single example image to describe their requirements, and the model platform can achieve an average top-4 identification accuracy of more than 80\%, indicating that recommending four models can satisfy users' needs in major cases on the benchmark dataset.  

\section{Problem Setup and Notions} 
In this paper, we explore a novel problem setting called \Setup, where users identify the most appropriate generative models for their specific purposes using one image. 
We assume there is a model platform, consisting of $M$ generative models $\left \{ f_m \right \}_{m=1}^M$. 
Each model is associated with a specification $S_m$ to describe its functionalities for future model identification. 
The platform consists of two stages: the submitting stage for model developers and the identification stage for users, respectively.

In the submitting stage, the model developer submits a generative model $f_m$ to the platform. 
Then, the platform assigns a specification $S_m$ to this model. Here, the specification $S_m = \cA_s \left (f_m, \PP \right )$ is generated by a specification algorithm $\mathcal{A}_s$ using the model $f_m$ and a prompt set $\PP = \left \{ \p_{k} \right \}_{k=1}^{N}$. 
If the model developer can provide a specific prompt set for the uploaded model, the generated specification would be more precise in describing its functionalities. 
In the identification stage, the users identify models from the platform using only one image $\x_{\tau}$. When users upload an image $\x_{\tau}$ to describe their purposes, the platform automatically calculates the pseudo-prompt $\widehat{\p}_{\tau}$ and then generates requirements $R_{\tau} = \mathcal{A}_r(\x_{\tau}, \widehat{\p}_{\tau})$ using a requirement algorithm $\cA_r$. 
Users can optionally provide corresponding prompt $\p_{\tau}$, setting $\widehat{\p}_r=\p_{\tau}$, to more precisely describe their purposes. 
During the identification process, the platform matches requirement $R_{\tau}$ with model specifications $\left \{ S_m \right \}_{m=1}^M$ using a evaluation algorithm $\cA_e$ and compute similarity score $\widehat{s}_{\tau, m} = \cA_e(S_m, R_{\tau})$ for each model $f_m$.
Finally, the platform returns the best-matched model with the maximum similarity score or a list of models sorted by $\left \{ \widehat{s}_{\tau, m} \right \}_{m=1}^M$ in descending order.

There are two main challenges for addressing \Setup\ setting: 
1) In the submitting stage, how to design $\cA_{s}$ to fully characterize the generative models for identification? 
2) In the identification stage, how to design $\cA_{r}$ and $\cA_{e}$ to effectively identify the most appropriate generative models for user needs?

\section{Proposed Method}
\label{sec:method}

In this section, we present our solution for the \Setup\ setting. 
Due to space limitations, we explain the RKME framework and its failure in \Setup\ in the \autoref{sec:analysis}. 
Our solution adopts a novel weighted term to capture the relationship between images and prompts in RKME, thereby enabling the model to be more precise in describing the functionalities of generative models. 
However, there are two issues remain: 
1) High dimensionality of images brings intractability of efficiency and similarity measurement; 
2) Cross-modality issue causes difficulties in calculating weight. 
To address these challenges, we employ a large pre-trained vision model $\cG(\cdot)$ to map images from image space to a common feature space. 
Subsequently, an image interrogator $\cI(\cdot)$ is adopted to convert $\x_{\tau}$ to corresponding pseudo prompt $\widehat{\p}_{\tau}$, thereby mitigating the cross-modality issues. 
Consequently, the similarity in the common feature space can be computed with the help of a large pre-trained language model $\cT(\cdot)$. 
We provide a detailed description of our proposal as follows. 

\paragraph{Submitting Stage}
The algorithm $\cA_s$ first samples images from the generative model $f_m$ using the prompt set: $\X_m = \left \{f_m(\p) | \p \in \PP \right \}$. The developer can optionally replace $\PP$ with a specific prompt set to generate a more precise specification. 
Then, the large pre-trained vision model $\cG(\cdot)$ is adopted to encode $\X_m$ as follows. The obtained feature representation $\Z_m$ is efficient and robust to compute the similarity between images, i.e., $\Z_m = \left \{\cG(\x) | \x \in \X_m \right \}$. 
Subsequently, $\cA_s$ encodes prompt set $\PP$ to the common feature representation using $\cT(\cdot)$: $\Q_m = \left \{ \cT(\p) | \p \in \PP \right \}$.
Finally, the specification $S_m$ of generative model $f_m$ is defined as follows: $S_m = \cA_s(f_m; \PP_m) = \left \{ \Z_m; \Q_m \right \}$.
Note that $S_m$ is automatically computed inside the platform, which is very convenient for developers to use and deduce their burden of uploading models. 
Additionally, the specification does not occupy a large amount of storage space on the platform since the only feature representation is storage. 

\paragraph{Identification Stage}
The users upload one single image $\x_{\tau}$ to describe their requirements and the platform describes the requirements with $R_{\tau}$ from $\x_{\tau}$.
Specifically, the requirement algorithm $\cA_r$ first generates feature representations of $\x_{\tau}$ using $\cG(\cdot)$, i.e., $\z_{\tau} = \cG(\x_{\tau})$. 
Subsequently, the pseudo-prompt $\widehat{\p}_{\tau}$ is generated by $\cI(\cdot)$, i.e., $\widehat{\p}_{\tau}=\cI(\x_{\tau})$, and converted to feature representations using $\cT(\cdot)$, i.e., $\widehat{\q}_{\tau} = \cT(\widehat{\p}_{\tau})$. 
The user can optionally replace $\widehat{\p}_{\tau}$ with a prompt $\p_{\tau}$ built on his understanding to precisely describe the requirement. 
Finally, the requirement is:$R_{\tau} = \cA_r(\x) = \left \{\z_{\tau}; \widehat{\q}_{\tau} \right \}$. 
Note that $R_{\tau}$ is automatically computed inside the platform, which is very easy to use for users. 
After the platform generates the requirement $R_{\tau}$, it will calculates the similarity score for each model $f_m$ using evaluation algorithm $\cA_e$:
\begin{equation}
    \cA_{e}(S_m, R_{\tau}) = \left \| \sum\limits_{i=1}^{N_m}\frac{1}{N_m} \frac{\widehat{\q}_{m, i} \widehat{\q}_{\tau}}{\|\widehat{\q}_{m, i}\| \|\widehat{\q}_{\tau}\|} k(\z_{m, i}, \cdot) - k(\z_{\tau}, \cdot) \right \|_{\cH_k}^2
     \label{eq:score-proposal}
\end{equation}
where the weighted term is defined as the cosine similarity between platform prompts $\widehat{\q}_{m, i} \in \widehat{\Q}_m$ and pseudo-prompt $\widehat{\q}_{\tau}$. $\W_m$ encodes the structure information of $\x_{\tau}$ within $\PP_m$ during the identification, which successfully captures the relation between images and prompts. The platform returns a list of models sorted in increasing order of similarity score obtained by \autoref{eq:score-proposal}.

\subsection{Discussion}
It is evident that our proposal for the \Setup\ scenario achieves a higher level of accuracy and efficiency when compared to model search techniques employed by existing model hubs.
For accuracy, our proposal elucidates the functionalities of generated models by capturing both the distribution of generated images and prompts, which allows for more accurate identification compared to the traditional model search method that relies on download ranks. 
For efficiency, our proposal achieves $O(T_r + M T_s)$ time for one identification, where generating requirement costs $T_r$ time and calculating similarity score costs $T_s$ time.  Moreover, with accurate identification results, users can save the efforts of browsing and selecting models, as well as reducing the consumption of network and computing. Additionally, our approach also has the potential to achieve further acceleration through the use of a vector database~\cite{LW-Guo23} such as Faiss~\cite{Faiss-johnson2019billion}. 

\section{Experiments}

In this section, we briefly introduce the experiment settings and main results. Detailed information about experiments is additionally provided in \autoref{sec:experiment}. 

\paragraph{Settings} We conduct experiments on a benchmark dataset described in \autoref{sec:dataset}. 
Our proposal is compared with three baseline methods: 
1) Download: The model is ranked based on the download volume~\cite{GPT-abs-2303-17580}, representing methods that ignore model capabilities. 
2) RKME-Basic: The model is identified using the basic RKME paradigm~\cite{LW-Guo23,LW-TanT0Z23}.
3) RKME-CLIP: The model is identified based on the combination of the RKME paradigm and CLIP model~\cite{CLIP-RadfordKHRGASAM21}. 
Two metrics, i.e., accuracy and rank, are adopted for evaluation. 
Accuracy evaluates the ability of methods to identify the most appropriate model, the higher the better. 
Rank evaluates the user's efforts in identifying the most appropriate models, the lower the better. 
Additionally, $Top$-$k$ accuracy is reported to indicate how many attempts can users find their satisfied models in major cases. 

\begin{figure}[t]
  \begin{minipage}{0.28\textwidth}
    \centering
    \includegraphics[width=\textwidth]{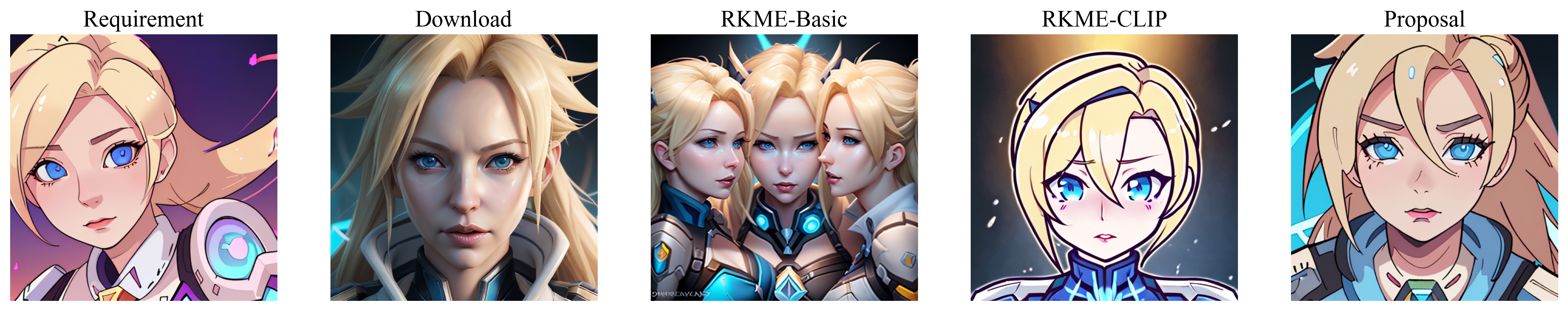}
    \includegraphics[width=\textwidth]{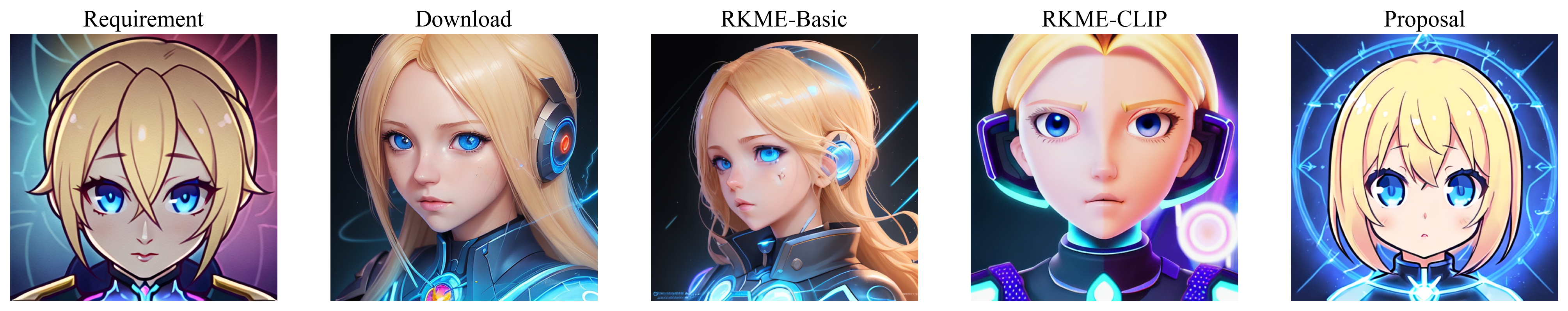}
    \includegraphics[width=\textwidth]{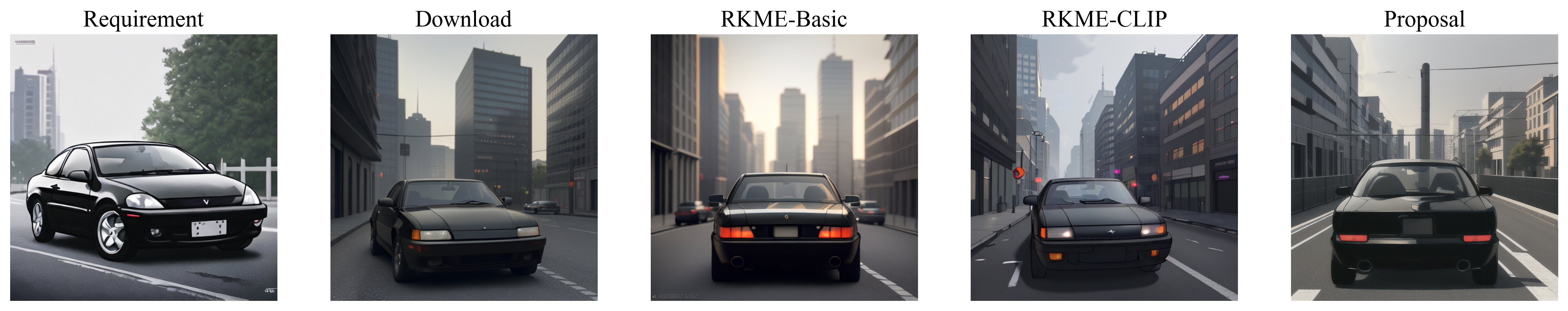}
    \includegraphics[width=\textwidth]{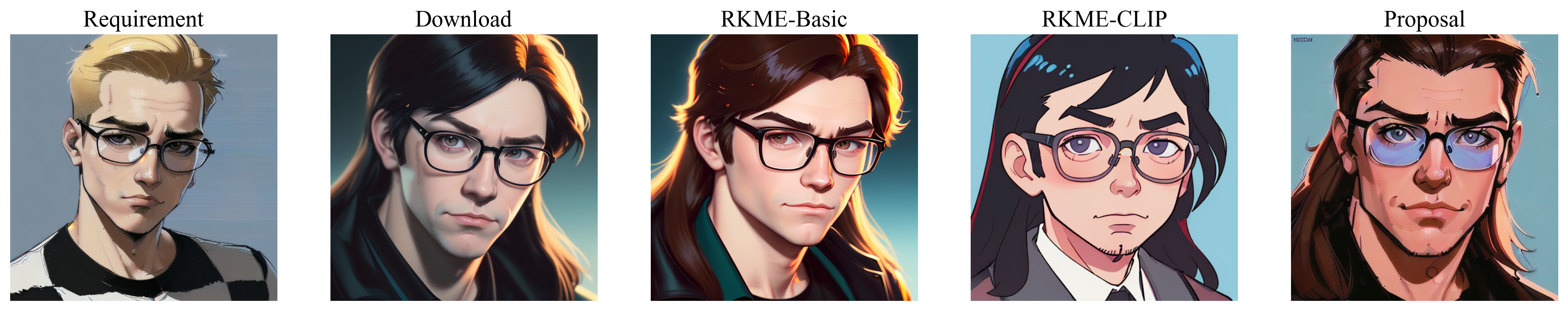}
    \caption{Visualization of generated images. }
    \label{fig:examples}
  \end{minipage}
  \hfill
  \begin{minipage}{0.70\linewidth}
    \captionof{table}{Performance of each method evaluated by $Top$-$k$ accuracy. 
  The results show that our proposal achieves 80\% top-4 accuracy, indicating that user only requires four models to satisfy their needs in major cases. }
    \label{tab:topk}
    \begin{center}
        \resizebox{\linewidth}{!}{
  \begin{tabular}{lrrrr}
  \toprule
      Methods &  Top-1 Acc. &  Top-2 Acc. &  Top-3 Acc. &  Top-4 Acc. \\
  \midrule
  Download &  0.062 &  0.125 &  0.188 &  0.250 \\
  RKME-Basic &  0.062 &  0.125 &  0.188 &  0.250   \\
  RKME-CLIP &  0.419 &  0.576 &  0.688 &  0.770  \\
  \midrule
  Proposal & \textbf{ 0.455} &  \textbf{0.614} &  \textbf{0.734} &  \textbf{0.812} \\
  \bottomrule
  \end{tabular}
  }
  \end{center}
  \end{minipage}
\end{figure}

\paragraph{Empirical Results}
As shown in \autoref{fig:performance}, our proposal achieves the best performance in both average accuracy and average rank, which demonstrates the effectiveness of our proposal. 
The Download and RKME-Basic methods cannot work in our setting because they do not consider the challenges of \Setup. The performance of the RKME-CLIP method improves significantly, indicating that the CLIP model can address the high dimensionality issue. Our proposal captures the relation between images and prompts, thereby giving the best performance.  
Table \ref{tab:topk} presents the results of $Top$-$k$ accuracy. These results show that our proposal achieves 80\% top-4 accuracy on the benchmark dataset, indicating that user only requires four attempts to satisfy their needs in major cases using our proposal to identify generative models. 
Finally, we show the visualization in \autoref{fig:examples}. 
The requirements are shown in the first column, and the generated images of each method using pseudo-prompts are shown in the remaining columns. Our proposal gives the most similar images.

\section{Conclusion}
In this paper, for the first time, we propose a novel problem called \emph{Generative Model Identification}. The objective of \Setup\ is to describe the functionalities of generative models precisely and enable the model to be accurately and efficiently identified in the future by users’ requirements. To this end, we present a systematic solution including a weighted RKME framework to capture the generated image distributions and the relationship between images and prompts, a large pre-trained vision-language model aimed at addressing dimensionality challenges, and an image interrogator designed to tackle cross-modality issues. Moreover, we built and released a benchmark platform based on stable diffusion models for \Setup. Extensive experiment results on the benchmark clearly demonstrate the effectiveness of our proposal. For example, our proposal achieves more than 80\% top-4 identification accuracy using just one example image to describe the users' requirements, indicating that users can efficiently identify the best-matched model within four attempts in major cases. 

In future work, we will endeavor to develop a novel generative model platform based on the techniques presented in this paper, aiming to provide a more precise description of generative model functionalities and user requirements. This will assist users in efficiently discovering models that align with their specific requirements. We believe this could facilitate the development and widespread usage of generative models.

\medskip
{
\small
\bibliographystyle{plainnat}
\bibliography{ref}
}


\newpage
\appendix
\section{Related Work}

Generative modeling~\citep{jebara2012machine} is a field of machine learning that focuses on learning the underlying distribution and generation of new samples for corresponding distribution. Recently, significant progress has been made in image generation with various methods. 
Generative Adversarial Networks (GANs)~\citep{GAN-ArjovskyCB17, GAN-BrockDS19, GAN-ChoiUYH20, GAN-GoodfellowPMXWOCB14} apply an adversarial approach to learn the data distribution. It consists of a generator and a discriminator playing a min-max game during the training process. 
Variational Autoencoders (VAEs)~\citep{VAE-KingmaW13, VAE-VahdatK20, VAE-OordVK17} is a variant of Auto-Encoder~(AE)~\citep{AE-WangYZ16}, where both consist of the encoder and decoder networks. The encoder in AE learns to map an image into a latent representation. Then, the decoder aims to reconstruct the image from that latent representation. 
Diffusion Models (DMs)~\citep{DM-NicholD21, SD-DhariwalN21, SD-Rombach22} leverages the concept of the diffusion process, consisting of forward and reverse diffusion processes. Noise is added to an image during the forward process and the diffusion model learns to denoise and reconstruct the image. 
With the development of the generative model, various generative model hubs/pools, e.g., HuggingFace, Civitai, have been developed. However, they lack model management and identification mechanisms, resulting in inefficiency for users to find the most suitable model. \citet{DBLP:journals/corr/abs-2210-03116} adopts a contrastive learning method to explore the search for deep generative models in terms of their contents.

Assessing the transferability of pre-trained models is related to the problem studied in this paper. 
Negative Conditional Entropy (NCE)~\citep{tran2019transferability} proposed an information-theoretic quantity~\citep{cover1999elements} to study the transferability and hardness between classification tasks. 
LEEP~\citep{nguyen2020leep} is primarily developed with a focus on supervised pre-trained models transferred to classification tasks. \citet{you2021logme} designs a general algorithm, which is applicable to vast transfer learning settings with supervised and unsupervised pre-trained models, downstream tasks, and modalities. However, these methods are not suitable for our \Setup\ problem because they impose significant computational overhead in terms of model inference during the identification process. Learnware~\citep{Lw-Zhou16} presents a general and realistic paradigm by assigning a specification to models to describe their functionalities and utilities, making it convenient for users to identify the most suitable models. Model specification is the key to the learnware paradigm. Recent studies~\citep{tan2022towards} are designed on Reduced Kernel Mean Embedding~(RKME)~\citep{LW-Wu23}, which aims to map the training data distributions to points in Reproducing Kernel Hilbert Space~(RKHS), and achieves model identification by comparing similarities in the RHKS. Subsequently, \citet{LW-Guo23} improves existing RKME specifications for heterogeneous label spaces. \citet{LW-TanT0Z23, tan2022towards} make their efforts to solve heterogeneous feature spaces. However, these studies primarily focus on classification tasks, overlooking the relationship between images and prompts, which is crucial for identifying generative models. Therefore, existing techniques are inadequate for addressing the \Setup\ problem, underscoring the pressing need for the development of new technologies specifically tailored to generative models.

\section{Problem Analysis}
\label{sec:analysis}

\subsection{Reduced Kernel Mean Embedding. }
\label{sec:baseline}
A baseline method to describe the model's functionality is the RKME techniques~\citep{LW-Wu23}. 
It maps data distribution of each model $f_m$ as corresponding specification $ S^{ \text{RKME} }_m = \left \{\x^{\text{RKME}}_{m, i} \right \}_{i=1}^{N^{\text{RKME}}_m}$, where $N^{\text{RKME}}_m$ is the reduced set size of $f_m$.
For one query image $\x_{\tau}$ from the users, the baseline method defines the requirement as $R_{\tau}^{\text{RKME}} = \{ \x_{\tau} \}$. 
Finally, the platform computes the similarity score in RKHS $\cH_k$ using evaluation algorithm $\cA_e^{\text{RKME}}$: 
\begin{equation}
    \cA_e^{ \text{RKME} }(S^{\text{RKME}}_m, R^{\text{RKME}}_{\tau}) 
    = \left \| \sum\limits_{i=1}^{ N^{\text{RKME}}_m }\frac{1}{N^{\text{RKME}}_m} k(\x^{\text{RKME}}_{m, i}, \cdot) - k(\x_{\tau} \cdot) \right \|_{\cH_k}^2
     \label{eq:spec-RKME}
\end{equation}
where $k(\cdot, \cdot)$ is the reproducing kernels associated with RKHS $\cH_k$.
This baseline method fails to capture the interplay between generated images $\X_m$ and the prompt set $\PP$, which is the probability distribution $p_{\theta_m}(\x_{0:T} | \p)$ inside the generative model $f_m$.  
We present an example to show this interplay is important otherwise the specification cannot distinguish two models in specific cases, resulting in unsatisfactory identification results. 

\begin{exmp}
Suppose that there are two simplified generative models $f_1$ and $f_2$ on the platform. 
$f_1$ generates scatter points following $x=\cos{(\text{p} \pi)}, y=\sin{(\text{p} \pi)}$. 
$f_2$ generates scatter points following $x=\sin{(\text{p} \pi)}, y=\cos{(\text{p} \pi)}$. 
The prompt set $\p$ follows $\mathcal{U}(-1, 1)$. 
The user wants to deploy the identified model conditioned on prompts $\p_{\tau}$ following distribution $\mathcal{U}(0.5, 0)$. 
In Figure~\ref{fig:example}, we show that the baseline method in \autoref{eq:spec-RKME} fails to distinguish two models $f_1$ and $f_2$ for the user. However, the two models function differently with $\p_{\tau}$. 
\begin{figure}[h]
    \centering
    \begin{subfigure}{0.25\textwidth}
        \centering
        \includegraphics[width=\textwidth]{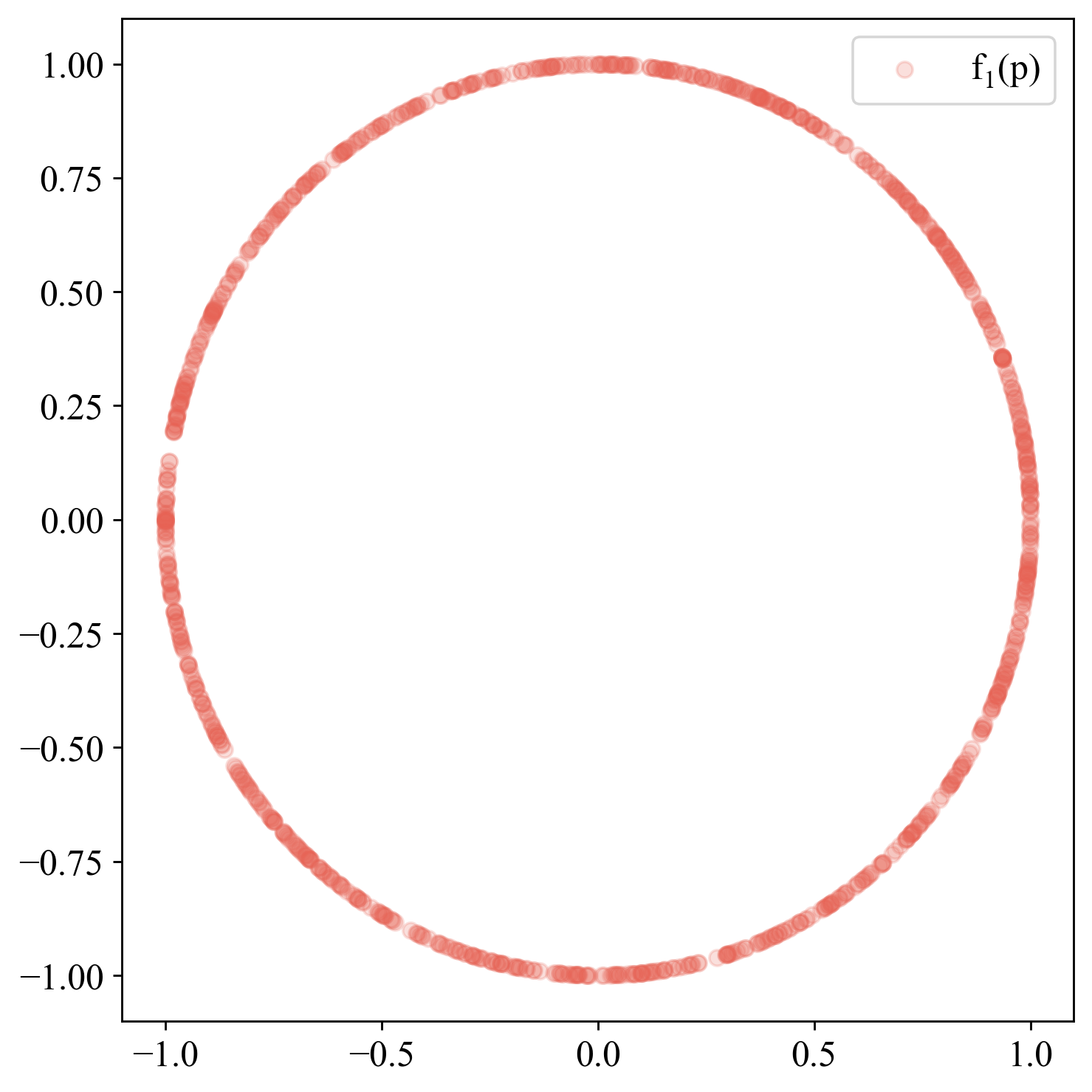}
        \caption{Distribution of specification $\X_1 \sim f_1(\p)$}
        \label{fig:exam-dist1}
    \end{subfigure}
    \quad
    \begin{subfigure}{0.25\textwidth}
        \centering
        \includegraphics[width=\textwidth]{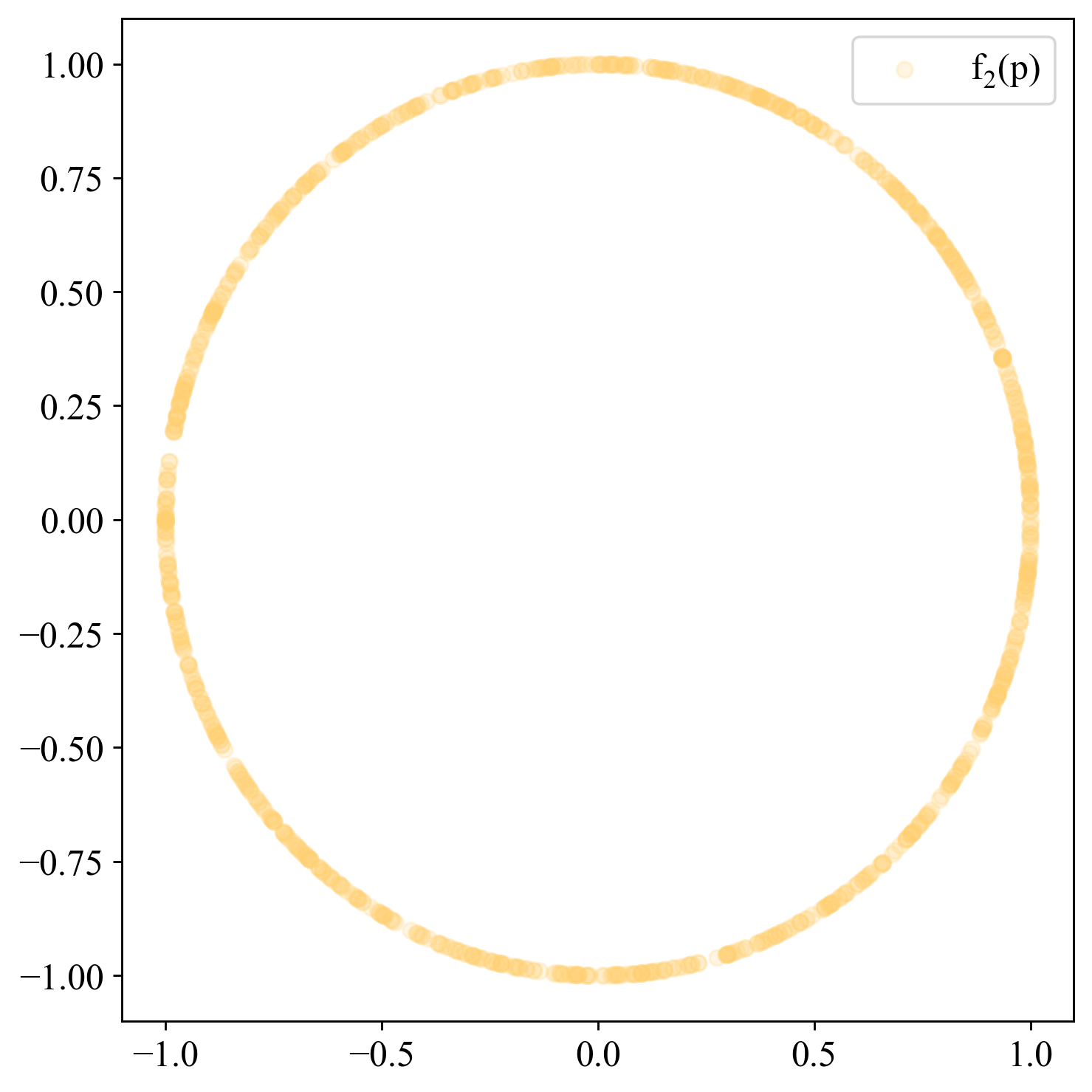}
        \caption{Distribution of specification $\X_2 \sim f_2(\p)$}
        \label{fig:exam-dist2}
    \end{subfigure}
    \quad
    \begin{subfigure}{0.25\textwidth}
        \centering
        \includegraphics[width=\textwidth]{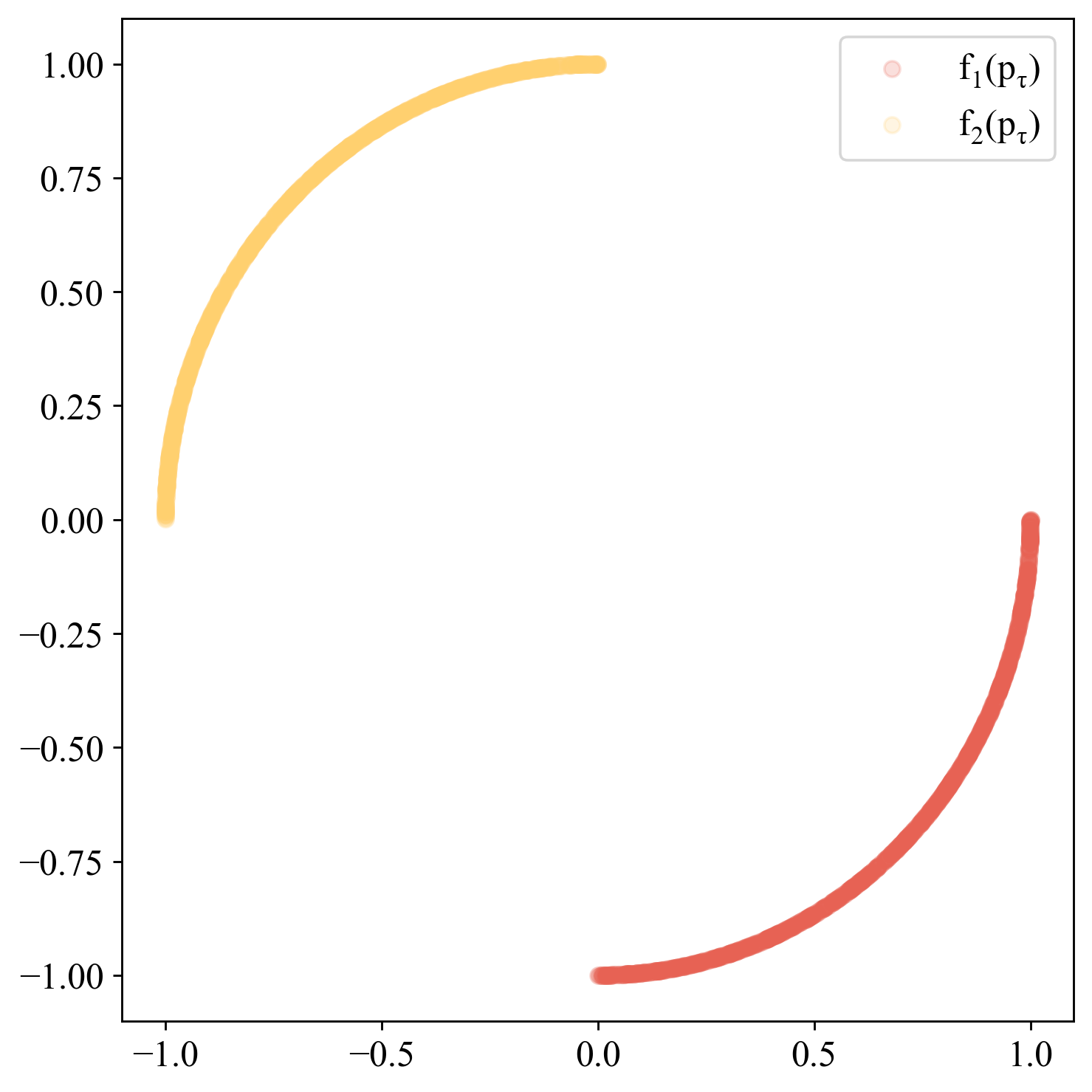}
        \caption{Distributions of $f_1(\p_{\tau})$ and $f_2(\p_{\tau})$}
        \label{fig:exam-func}
    \end{subfigure}
    \caption{Baseline method in \autoref{eq:spec-RKME} fails to distinguish two different models for users. }
    \label{fig:example}
\end{figure}
\autoref{fig:exam-dist1} and \autoref{fig:exam-dist2} show that although models $f_1$ and $f_2$ function differently, the data distribution $\X_1 \sim f_1(\p)$ and $\X_2 \sim f_2(\p)$, conditioned on the default prompt distribution $\p$, could be identical.
Therefore, the specificaions $S_1^{\text{RKME}}$ and $S_2^{\text{RKME}}$ are identical, resulting in the same similarity scores $\cA_e^{\text{RKME}}(S^{\text{RKME}}_1, R^{\text{RKME}}_{\tau})$ and $\cA_e^{\text{RKME}}(S^{\text{RKME}}_2, R^{\text{RKME}}_{\tau})$.
However, \autoref{fig:exam-func} shows that two models $f_1$ and $f_2$ generate different data distributions $f_1(\p_{\tau})$ and $f_2(\p_{\tau})$ conditioned on the user prompt distribution $\p_{\tau}$. 
\label{exam:spec-RKME}
\end{exmp}
\begin{remark}
Example~\ref{exam:spec-RKME} shows us that overlooking the interplay between images and prompts leads to impossible cases for distinguishing generative models effectively. 
Existing RKME studies mainly focus on classification tasks, which can implicitly model the tasks through data distribution since the class space is discrete and small. For generative models, we have to explicitly model the model's functionality, i.e., the relation between images and prompts, to achieve satisfied identification results. 
\end{remark}

\subsection{Weighted RKME Framework}
\label{sec:WRKME}
Motivated by our analysis, how to incorporate the relationship between images and prompts in model specification and identifying process is the key challenge for our \Setup\ setting. 
Inspired by existing studies~\citep{MMD-LiSZ15, MMD-RenZLL16} about the conditional maximum mean discrepancy, we propose to consider the above relation using a weighted formulation of \autoref{eq:spec-RKME}:
\begin{equation}
    \cA^{\text{Weighted}}_{e}(S^{\text{Weighted}}_m, R^{\text{Weighted}}_{\tau}) = \left \| \sum\limits_{i=1}^{N_m}\frac{1}{N_m} w_{m, i} \cdot k(\x_{m, i}, \cdot) - k(\x_{\tau}, \cdot) \right \|_{\cH_k}^2
     \label{eq:spec-proposal}
\end{equation}
where $\W_m=\left \{ w_{m, i}\right \}_{i=1}^{N_m}$ are required to measure the relation between user image $\x_{\tau}$ and prompt set $\PP$. 
Here, we make the simplifications $R^{\text{Weighted}}_{\tau} = \x_{\tau}$ and $S^{\text{Weighted}}_{m} = \X_m$ in \autoref{eq:spec-proposal}. 
This raises challenges inherent in dimensionality since stable diffusion models produce high-quality images. 
Moreover, measuring the relation using $\W_m$ is also a challenging problem and encounters cross-modality issues. 

\section{Detailed Experiment Settings and Results}
\label{sec:experiment}

\subsection{Model Platform and Task Construction}
\label{sec:dataset}
In practice, we expect model developers to submit their models and corresponding prompts to the model platform. 
And we expect users to identify models for their real needs. 
In our experiments, we constructed a model platform and user identification tasks respectively to simulate the above situation. 
For the construction of the model platform, we manually collect $M=16$ different stable diffusion models $\{ f_1, \ldots, f_{M} \}$ from one popular model platform, CivitAI, as uploaded generative models on the platform. 
Note that these collected models belong to the same category to simulate the real process in which users first trigger category filters and then select the models. 
We construct 55 prompts $\{ \p_1, \ldots, \p_{55} \}$ as default prompt set $\PP$ of platform. 
For task construction, we construct 18 evaluation prompts $\{ \p_{\tau_1}, \ldots, \p_{\tau_{18}} \}$ for each model on the platform to generate testing images with random seed in $\{0, 1,2,3,4,5,6,7,8,9\}$, forming $N_{\tau} = 18\times 16\times10=2880$ different identification tasks $\left \{ ( \x_{\tau_i}, t_i ) \right \}_{i=1}^{N_{\tau}}$, where each testing image $\x_{\tau_i}$ is generated by model $f_{t_i}$ and its best matching model index is $t_i$. 
Here, we ensure that there is no overlap between $\{ \p_1, \ldots, \p_{55} \}$ and $\{ \p_{\tau_1}, \ldots, \p_{\tau_{18}} \}$ to ensure the correctness of the evaluation.

\subsection{Comparison Methods.} 
Initially, we compare it with the traditional model search method called Download.
This method is used to simulate how users search generative models according to their downloading volumes~\citep{GPT-abs-2303-17580}, where users will try models with high downloading volume first. 
This baseline method can represent a family of methods that employ statistical information without regard to model capabilities. 
We also consider the basic implementation of the RKME specification~\citep{LW-Wu23} as a baseline method RKME-Baisc for our \Setup\ problem. 
The details of generating specifications, and identifying models are presented in \autoref{sec:baseline}. 
Furthermore, we compare our proposed method with a variant of the basic RKME specification, that is, RKME-CLIP, which calculates specifications in the feature representation space encoded by the CLIP model~\citep{CLIP-RadfordKHRGASAM21}. 
The results obtained from RKME-CLIP further support our viewpoint on the critical challenges posed by dimensionality.

\subsection{Implementation Details.}
We adopt the official code in~\citet{LW-Wu23} to implement the RKME-Basic method and the official code in~\citet{CLIP-RadfordKHRGASAM21} to implement the CLIP model.
For RKME-Basic and RKME-CLIP methods, we follow the default hyperparameter setting of RKME in previous studies~\citep{LW-Guo23}. 
We set the size of the reduced set to 1 and choose the RBF kernel~\citep{RBF-XuKY94} for RKHS. The hyperparameter $\gamma$ for calculating RBF kernel and similarity score is tuned from $\left \{0.005, 0.006, 0.007, 0.008, 0.009, 0.01, 0.02, 0.03, 0.04, 0.05\right \}$ and set to $0.02$ in our experiments. Experiment results below show that our proposal is robust to $\gamma$.

\subsection{Ablation Study}
In order to comprehensively evaluate the effectiveness of our proposal, we investigate whether each component contributes to the final performance. 
We additionally compare our proposal with two variants, called RKME-CLIP and RKME-Concat. 
RKME-CLIP adopts the CLIP model to extract the feature representation for constructing RKME specifications. 
RKME-Concat adopts both vision and text branches of the CLIP model to extract representations of images and prompts. 
It combines two modes of representation for constructing RKME specifications. 
We report accuracy and rank metrics in \autoref{tab:ablation}.
The performance of RKME-CLIP demonstrates that employing large pre-trained models is an effective approach for addressing dimensionality issues.
The performance of RKME-Concat demonstrates the benefits of considering both images and prompts for model identification.
Our results achieve the best performance, and demonstrate the effectiveness of our weighted formulation in \autoref{eq:spec-proposal} and our specifically designed algorithm in \autoref{eq:score-proposal}. 
\begin{table}[t]
  \captionof{table}{Ablation study. For accuracy, the higher the better. For rank, the lower the better. The best performance is in bold. }
  \label{tab:ablation}
  \centering
\begin{tabular}{lrrr}
\toprule
Methods &  Acc. &  Top-2 Acc. &  Rank \\
\midrule
Download & 0.062 &       0.125 & 8.500 \\
RKME-Basic & 0.062 &       0.125 & 8.500 \\
RKME-CLIP & 0.419 &       0.576 & 3.130 \\
RKME-Concat & 0.433 &       0.602 & 2.938 \\
\midrule
Proposal & \textbf{0.455} &       \textbf{0.614} & \textbf{2.852} \\
\bottomrule
\end{tabular}
\end{table}

\subsection{Hyperparameter Robustness}
\begin{figure}[t]
  \centering
  \includegraphics[width=0.5\textwidth]{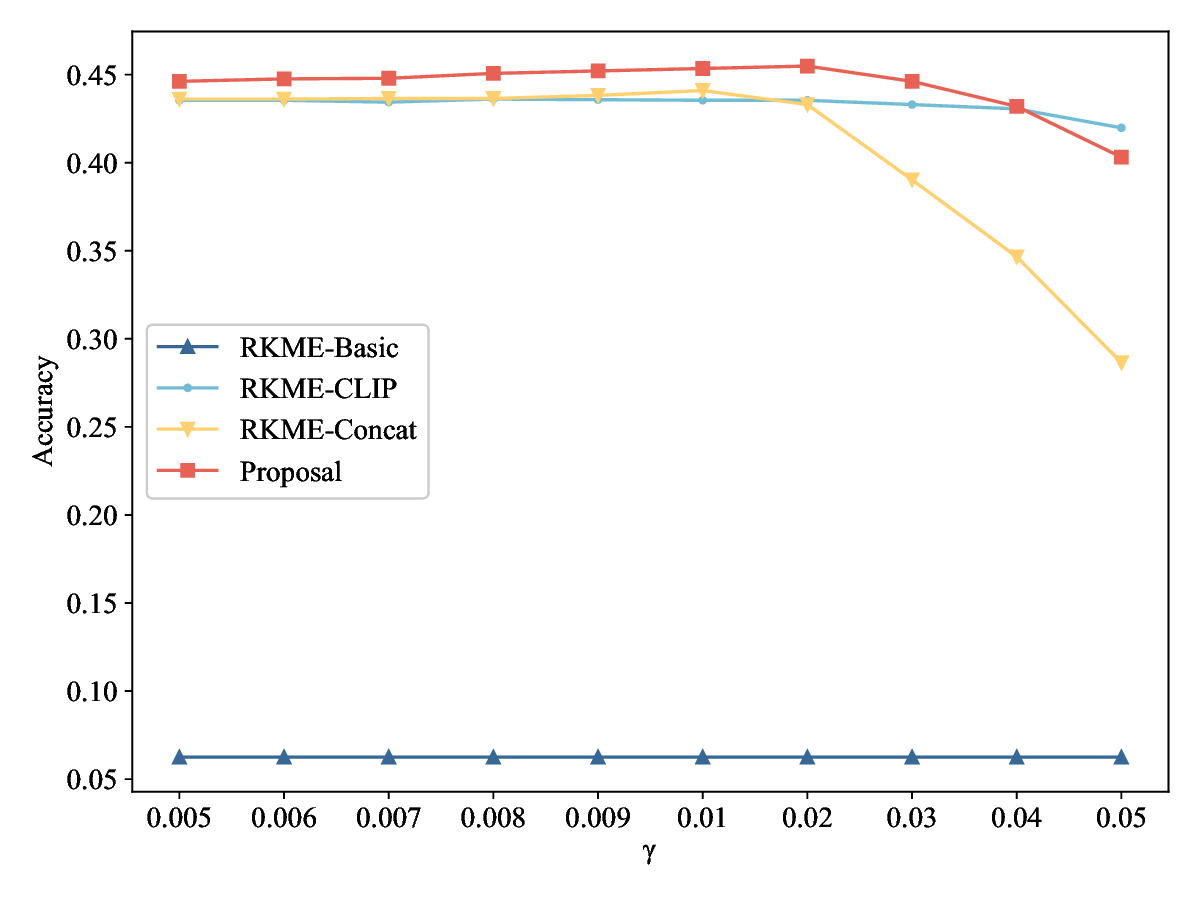}
  \caption{The accuracy with varying values of $\gamma$ was evaluated. 
  The results demonstrate that our proposal is robust to slight changes in the value of $\gamma$.}
  \label{fig:gamma}
\end{figure}

We evaluate the robustness of each method to the hyperparameter $\gamma$ in \autoref{fig:gamma}. 
The results demonstrate that our proposed method exhibits robust performance across a wide range of $\gamma$ values. 
However, as $\gamma$ continues to increase, the performance of both our proposal and the baseline methods begins to degrade. 
This observation highlights the importance of tuning the hyperparameter $\gamma$ before deploying our method in practical applications. 
Once $\gamma$ is properly tuned, our method can operate robustly due to its hyperparameter robustness within a broad range.

\end{document}